\pdfoutput=1

\documentclass[11pt]{article}

\usepackage{authblk}

\usepackage[]{acl}

\usepackage{times}
\usepackage{latexsym}

\usepackage[T1]{fontenc}

\usepackage[utf8]{inputenc}

\usepackage{microtype}

\usepackage{booktabs}
\usepackage{makecell}
\usepackage{stfloats}  
\usepackage{kotex}
\usepackage{algorithm}
\usepackage{algpseudocode}

\usepackage{amsmath}
\usepackage{amssymb}

\usepackage{cleveref}
\usepackage{graphicx}

\crefformat{section}{\S#2#1#3} 
\crefformat{subsection}{\S#2#1#3}
\crefformat{subsubsection}{\S#2#1#3}

\usepackage{listings}
\usepackage{xcolor}

\definecolor{codegreen}{rgb}{0,0.6,0}
\definecolor{codegray}{rgb}{0.5,0.5,0.5}
\definecolor{codepurple}{rgb}{0.58,0,0.82}
\definecolor{backcolour}{rgb}{0.95,0.95,0.92}

\lstdefinestyle{mystyle}{
    backgroundcolor=\color{backcolour},   
    commentstyle=\color{codegreen},
    keywordstyle=\color{magenta},
    numberstyle=\tiny\color{codegray},
    stringstyle=\color{codepurple},
    basicstyle=\ttfamily\footnotesize,
    breakatwhitespace=false,         
    breaklines=true,                 
    captionpos=b,                    
    keepspaces=true,                 
    numbersep=5pt,                  
    showspaces=false,                
    showstringspaces=false,
    showtabs=false,                  
    tabsize=2
}

\DeclareMathOperator*{\argmin}{arg\,min} 

\title{MCS-SQL: Leveraging Multiple Prompts and Multiple-Choice Selection For Text-to-SQL Generation}

\renewcommand*{\Affilfont}{\normalsize\normalfont}

\newsavebox\affbox
\author[]{Dongjun Lee}
\author[]{Choongwon Park}
\author[]{Jaehyuk Kim}
\author[]{Heesoo Park}
\affil[]{\Affilfont Dunamu}
\affil[]{\textit {\{tonny, elvie, loki, belle\}@dunamu.com}}

\begin{document}
\maketitle

\begin{abstract}
Recent advancements in large language models (LLMs) have enabled in-context learning (ICL)-based methods that significantly outperform fine-tuning approaches for text-to-SQL tasks. However, their performance is still considerably lower than that of human experts on benchmarks that include complex schemas and queries, such as BIRD.
This study considers the sensitivity of LLMs to the prompts and introduces a novel approach that leverages multiple prompts to explore a broader search space for possible answers and effectively aggregate them.
Specifically, we robustly refine the database schema through schema linking using multiple prompts. Thereafter, we generate various candidate SQL queries based on the refined schema and diverse prompts. 
Finally, the candidate queries are filtered based on their confidence scores, and the optimal query is obtained through a multiple-choice selection that is presented to the LLM.
When evaluated on the BIRD and Spider benchmarks, the proposed method achieved execution accuracies of 65.5\% and 89.6\%, respectively, significantly outperforming previous ICL-based methods.
Moreover, we established a new SOTA performance on the BIRD in terms of both the accuracy and efficiency of the generated queries.

\end{abstract}

\begin{figure*}
\centering\includegraphics[scale=0.2]{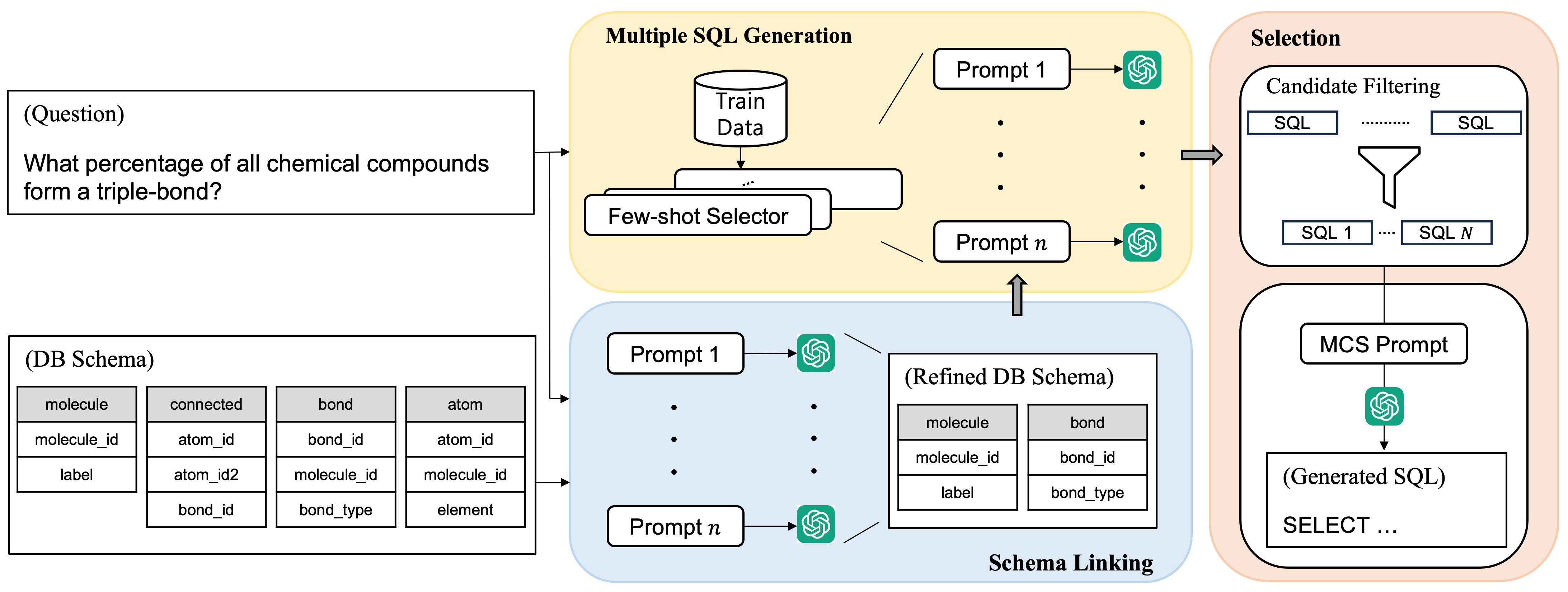}
\caption{\label{fig:methodology} Overview of the proposed methodology, including three steps: schema linking, multiple SQL generation, and selection. }

\end{figure*}

\section{Introduction}

The text-to-SQL task involves translating a natural language question into SQL and is crucial for natural language interfaces to databases (NLIDB) systems.
With the recent advancements in large language models (LLMs), in-context learning (ICL)-based approaches for text-to-SQL \citep{din-sql, dail-sql, exploring-cot} have demonstrated significant performance improvements over traditional fine-tuning methods \citep{hui2022s2sql, schema-linking-4, schema-linking-5, li2023graphix}. 
Notably, \citet{evaluating} showed that these methods even surpassed gold reference queries in terms of human evaluation within the Spider benchmark.
However, for the more challenging BIRD \citep{bird} benchmark, characterized by its complex database (DB) schemas and queries, the accuracies of ICL-based methods have not exceeded 60\%, which is significantly lower than the 93.0\% achieved by humans. This gap underscores the need for further advancements in the ICL approach to serve as an NLIDB system.

A significant limitation of LLMs across various tasks is their sensitivity to the structure and content of prompts. 
Even for semantically identical prompts, LLMs can generate drastically varying responses because of factors such as the order of sentences \citep{consistency-analysis, primacy-effect}, choice of demonstration examples \citep{liu2022makes, medprompt}, and the sequence in which these examples are presented \citep{fantastically-ordered-prompts}. Our experiments confirmed a similar tendency in the text-to-SQL context, where alterations in the schema presentation (\cref{sec:effect-schema-linking}) and the choice of few-shot examples (\cref{sec:effect-few-shot}) resulted in variations in LLM outputs.

In this study, to improve the accuracy and robustness of LLMs for text-to-SQL, we introduce a novel approach that leverages multiple prompts to generate various candidate answers and effectively aggregates them. We utilize the sensitivity of LLMs to prompts to explore a broader search space for answers using different prompts.
As shown in Figure~\ref{fig:methodology}, the SQL generation process comprises three steps: schema linking, multiple SQL generation, and selection. 
Initially, the schema-linking phase robustly selects tables and columns relevant to the question from the DB schema using multiple prompts.
Subsequently, the generation phase employs various prompts to produce diverse candidate SQL queries, ensuring a broader exploration of potential queries. Finally, the selection phase filters candidate queries based on confidence scores, and the optimal query is selected through multiple-choice selection (MCS) that is presented to the LLM.

We evaluated our methodology using two benchmarks, BIRD \citep{bird} and Spider \citep{spider}. For BIRD, we achieved an execution accuracy (EX) of 65.5\% and a valid efficiency score (VES) of 71.4\%, which outperforms the previous state-of-the-art (SOTA) ICL-based method \citep{mac-sql} by +5.9\% and +3.7\%, respectively, setting a new SOTA performance for the BIRD. 
In addition, for the Spider test set, we achieved an EX of 89.6\%, which outperforms the existing SOTA ICL-based approach \citep{dail-sql} by 3.0\%.

\section{Related Work}

\paragraph{Prompt Engineering}
Prompt engineering, which is the study of designing effective prompts, is an active research area as prompts significantly impact the performance of LLMs across various NLP tasks.
A prominent example is the chain-of-thought (CoT) prompting \citep{cot}, which employs manually crafted examples to guide the LLM to generate intermediate reasoning steps prior to deriving the answer. This technique has demonstrated significant performance enhancements across various tasks. \citet{cot-zero-shot} further demonstrated that LLMs can explain their reasoning steps even in the absence of human-annotated examples.

In addition to CoT, self-consistency decoding \citep{self-consistency} has been proposed, wherein multiple answers are sampled from the LLM before selecting one through a majority vote. 
Compared with greedy decoding, this technique facilitates the exploration of various reasoning paths and has exhibited substantial performance gains. For more effective explorations of the reasoning steps, variations such as tree-of-thought \citep{cot} and graph-of-thought \citep{got} have also been proposed.

However, because these approaches rely on a single prompt to generate the reasoning steps, we hypothesize that they fail to mitigate the issue of LLMs' high sensitivity to prompts, thereby exploring only a limited search space. 
Several studies have shown that even when presented with semantically identical prompts, the outputs of LLMs vary based on factors such as sentence structure \citep{webson2022prompt}, sequence of sentences \citep{consistency-analysis, primacy-effect, pezeshkpour2023large}, choice of few-shot examples \citep{liu2022makes, medprompt}, and the order in which the examples are presented \citep{fantastically-ordered-prompts}.
Therefore, we use multiple distinct prompts for a wider exploration of potential answers, thereby mitigating the LLM's sensitivity issue of LLMs to prompts and ensuring the generation of more robust answers.

\paragraph{ICL for Text-to-SQL}
As ICL-based approaches have shown remarkable performance in text-to-SQL tasks, various studies have focused on creating better prompts for text-to-SQL.
Several studies have focused on applying prompting techniques such as CoT or least-to-most \citep{least-to-most} for text-to-SQL generation \citep{din-sql, exploring-cot, mac-sql}.
However, these methods rely on fixed sets of manually crafted examples, and their performance can vary significantly depending on the selection of these examples.
In this work, instead of relying on fixed human-labeled examples, we dynamically select few-shot examples from the training data based on the test sample. Some studies have aimed to determine a more effective few-shot selection strategy for text-to-SQL \citep{prompting-gpt-35, enhancing-text-to-sql, dail-sql}. However, unlike these studies, which focused on determining a single optimal selection strategy, we employ a parallel approach that employs various few-shot selection strategies with multiple prompts and effectively aggregates them.

\section{Methodology}
As shown in Figure~\ref{fig:methodology}, the proposed method comprises three steps: (1) schema linking, wherein tables and columns irrelevant to the question from the DB schema are excluded; (2) multiple SQL generation, wherein multiple candidate SQL queries are generated based on various prompts; and (3) selection, wherein the most accurate SQL query is selected from among the candidates.

\subsection{Schema Linking}

Schema linking involves identifying relevant tables and columns from a DB to convert a natural language question into an SQL query \citep{schema-linking-1}.
The introduction of schema linking has significantly improved the performances of both fine-tuning-based \citep{schema-linking-2, schema-linking-3, schema-linking-4, schema-linking-5} and ICL-based \citep{c3-sql, din-sql, mac-sql} approaches.

We perform schema linking in two steps: first, the tables related to the natural language query are extracted (table linking). Thereafter, the necessary columns within those tables are extracted (column linking). 
We employ multiple prompts in both phases with the aim of achieving a high recall.

\subsubsection{Table Linking}

In table linking, the DB schema and question are input into the LLM, which extracts a list of reference tables to generate the SQL query. Inspired by zero-shot-CoT \citep{cot-zero-shot}, we ask the LLM to explain why each table is necessary, instead of just selecting a list of tables. To easily parse the LLM's answer, we ask it to respond in the JSON format, as described in Listing~\ref{lst:table-linking}.

\begin{lstlisting}[language={}, caption=Prompt template for table linking. An example of the full prompt is presented in Appendix~\ref{appendix:table-linking-prompt}., label={lst:table-linking}]
### For a given DB schema and question, extract the list of tables required to write the SQL query.

### DB schema: ...
### Question: ...

Your answer should be in the json format:
{
  "reasoning": "..." # The reason for selecting each table.
  "answer": [...] # List of selected tables.
}

### Your answer:
\end{lstlisting}

To enhance the robustness of table linking, we utilize multiple prompts. Various studies have demonstrated that LLM outputs are significantly affected by the sequence of input sentences \citep{consistency-analysis, primacy-effect, lost-in-the-middle}. Similarly, our experiments (\cref{sec:effect-schema-linking}) revealed that the output of schema-linking output of LLMs also depends on the sequence in which the tables and columns are arranged in the prompts.
To minimize the influence of the table order, we randomly shuffle the order of tables, generating $p_t$ distinct prompts. For each prompt, we obtain $n$ responses from the LLM by using a high sampling temperature. The final table-linking output is derived from the union of all responses, amounting to $p_t \cdot n$ table lists. We use a union operation because including unnecessary tables in table linking does not significantly impact the subsequent SQL-generation process; however, omitting the necessary tables prevents the generation of the correct SQL query.

\subsubsection{Column Linking}

For column linking, we ask the LLM to extract the columns required for converting a question into an SQL query using a prompt similar to that used in table linking.
The prompt includes only the schemas of the tables selected during table linking, instead of the entire DB schema.
Because the same column name can exist in different tables, we instruct the LLM to provide the answer in the [\texttt{table\_name}].[\texttt{column\_name}] format.

Similar to table linking, the order of the tables and columns is randomly shuffled to generate $p_c$ unique prompts. Subsequently, $n$ LLM responses are generated for each prompt, where each response represents a selected column list. The column-linking output is the union of all $p_c * n$ responses.

In the subsequent SQL-generation steps, when providing the DB schema to LLM, only the tables and columns selected through schema linking are provided instead of the full schema.

\subsection{Multiple SQL Generation}
To address the sensitivity of the LLM to prompts, we generate various SQL queries based on multiple, distinct prompts. Several studies have demonstrated that the output of an LLM can differ significantly depending on the few-shot examples provided \citep{liu2022makes, wu-etal-2023-self}, and even on the sequence in which these examples are presented \citep{fantastically-ordered-prompts, calibrate-before-use}. 
To effectively leverage this variability, we generate multiple prompts by varying both the selection method of the few-shot examples and the order in which they are presented, thereby ensuring a broader exploration of potential SQL queries.

\subsubsection{Few-Shot Examples Selection}
For each test sample, a set of few-shot examples is selected from the training dataset. To generate multiple prompts with different examples, we use two distinct selection strategies: one that leverages question similarity and another that utilizes masked question similarity.
In the question similarity-based approach, we select the top-k questions from the training dataset that have the nearest sentence embeddings to the natural language question of the test sample.

Similarly, the masked question similarity-based approach considers the embedding similarity of masked questions, wherein tokens specific to the DB schema in the question are masked. This masking allows determining the similarity of questions in terms of generating similar queries by disregarding schema-specific content.
We employ an LLM for the masking process by presenting it with the DB schema and question, and asking it to replace the table names, column names, and values with special tokens. 
The prompt for this question masking is presented in Appendix~\ref{appendix:question-masking-prompt}.

Through these two few-shot selection strategies, we generate $p_q$ different prompts, including one derived exclusively from question similarity, another solely from masked question similarity, and additional prompts created by integrating examples from both strategies in various sequences.

\subsubsection{SQL Generation}

\begin{lstlisting}[language={}, caption=Prompt template for SQL generation. An example of the full prompt is presented in Appendix~\ref{appendix:sql-generation-prompt}., label={lst:sql-generation}]
### Generate the correct SQL query for a given DB schema and question.

### Gold Examples:
- Question: ...
- Gold SQL: ...
...

### DB Schema: ...
### Sample Table Contents: ... 
### Question: ...

Your answer should be in the json format:
{
  "reasoning": ".." # The reasoning steps behind the generated SQL query
  "sql": ".." # The generated SQL query.
}

### Your answer: 
\end{lstlisting}
As illustrated in Listing~\ref{lst:sql-generation}, our SQL-generation prompt includes few-shot examples, a DB schema, sample table contents, and a natural language question. 
The few-shot examples comprise questions and their corresponding gold SQL pairs. To conserve the limited length of the prompt, we exclude the schema of the target DB for each question.
Regarding the DB schema, we selectively present only the tables and columns selected during the schema-linking process to avoid burdening the LLM with irrelevant information.
Additionally, we embed sample table contents in the CSV format within the prompt to facilitate the LLM's comprehension of potential values in each column, thereby providing practical insight into the data structure of the DB.
Finally, we instruct LLM not only to produce the SQL query but also to explain the reasoning behind its generation, thereby enhancing the model's interpretability and accuracy.

For each prompt, we generate $n$ responses from the LLM using a high sampling temperature, resulting in $p_q \cdot n$ candidate SQL queries being generated.

\subsection{Selection}
\label{subsec:selection}
The selection step aims to select the most accurate query from the candidate queries. Initially, the candidate pool is filtered based on confidence scores, and the LLM is then tasked with selecting the most accurate query from the refined pool.

\subsubsection{Candidate Filtering}

To select the most accurate query among the candidates, we first narrow down the candidate pool. Queries with the same execution results are grouped together, and only the fastest query from each group is retained. Additionally, queries with low confidence scores are excluded from the candidates.

In detail, all candidate queries are executed on the DB, and queries that result in syntax errors or timeouts are removed from the candidate pool.
Next, the confidence score for each query in the candidate pool is calculated, which is determined based on the number of queries that produce the same execution result. Formally, for a candidate pool $C$ containing $N$ queries $\{q_1, \ldots, q_N\}$, the confidence of query $q_i$ is computed as follows:
\begin{gather}
\text{confidence}(q_i) = \frac{1}{N} \sum_{j=1}^{N} [\text{exec}(q_j) = \text{exec}(q_i)]
\end{gather}
where $\text{exec}(q_i)$ denotes the execution result for query $q_i$. 

Among the queries in candidate pool $C$ with identical execution results, only the query with the minimum execution time is selected as follows:
\begin{gather}
C' = \bigcup_{R \in \mathcal{R}(C)} \argmin_{q_i \in C, \text{exec}(q_i) = R} {\text{exec\_time}(q_i)}
\end{gather}
where $\mathcal{R}(C)$ represents the set of all unique execution results from all queries in $C$, and $\text{exec\_time}(q_i)$ denotes the execution time for query $q_i$.

Finally, all queries in $C'$ with confidence score below threshold $T$ are excluded as follows:
\begin{gather}
\label{eq:confidence-threshold}
C'' = \{ q_i \in C' \mid \text{confidence}(q_i) \geq T\},
\end{gather}
resulting in a refined candidate pool $C''$.

\subsubsection{Multiple-Choice Selection (MCS)}
Following the filtering process, we utilize the LLM to select the most accurate query among the candidates through a multiple-choice question.

\begin{lstlisting}[language={}, caption=Prompt template for SQL selection. An example of the full prompt is presented in Appendix~\ref{appendix:mcs-prompt}., label={lst:mcq-prompt}]
### For a given DB schema and question, select the most accurate query among the candidate SQL queries.

### DB schema: ...
### Question: ...
### Candidate SQLs:
1. SQL1
2. SQL2
3. SQL3

Your answer should be in the json format:
{
  "reasoning": ".." # The reasoning steps for selecting the correct SQL query.
  "sql": ".." # The selected SQL query.
}

### Your answer: 
\end{lstlisting}

As shown in the Listing~\ref{lst:mcq-prompt}, we present a set of candidate SQL queries to the LLM and request it to select the most accurate query for a given DB schema and question.
Candidate queries are provided in descending order of confidence scores, considering the tendency of LLMs to favor options that appear earlier in the multiple-choice questions \citep{primacy-effect, zheng2023large}. 
The LLM is required to not only select an SQL query but also provide the reasons for its selection.
We sample $n$ responses from the LLM and determine the final SQL query through a majority vote.

\begin{table*}
    \centering
    \begin{tabular}{lcc|cc}
    \toprule
        & \multicolumn{2}{c}{Dev} & \multicolumn{2}{c}{Test} \\
        Model & EX & VES & EX & VES \\
    \midrule
    GPT-4 (zero-shot) & 46.4 & 49.8 & 54.9 & 60.8 \\
    DIN-SQL + GPT-4 \citep{din-sql} & 50.7 & 58.8 & 55.9 & 59.4 \\
    DAIL-SQL + GPT-4 \citep{dail-sql} & 54.8 & 56.1 & 57.4 & 62.0 \\
    MAC-SQL + GPT-4 \citep{mac-sql} & 57.7 & 58.8 & 59.6 & 67.7 \\
    MCS-SQL + GPT-4 (Ours) & \textbf{63.4} & \textbf{64.8} & \textbf{65.5} & \textbf{71.4} \\
    \bottomrule
    \end{tabular}
    \caption{Execution accuracies (EX) and valid efficiency scores (VES) for the BIRD dev and test sets.}
    \label{table:bird-main}
\end{table*}

\begin{table}
\centering
\begin{tabular}{lcc}
    \toprule
        Model & Dev & Test \\
    \midrule
        \makecell{GPT-4 (zero-shot)} & 74.6 & - \\
        \makecell{DIN-SQL + GPT-4 \\ \citep{din-sql}} & 82.8 & 85.3 \\
        \makecell{DAIL-SQL + GPT-4 \\ \citep{dail-sql}} & 84.4 & 86.6 \\
        \makecell{MAC-SQL + GPT-4 \\ \citep{mac-sql}} & 86.8 & - \\
        \makecell{MCS-SQL + GPT-4 (Ours)} & \textbf{89.5} & \textbf{89.6} \\
    \bottomrule
\end{tabular}
\caption{Execution accuracies for the Spider dev and test sets. "-" denotes that the model did not report performance for the test set.}
\label{table:spider-main}
\end{table}

\section{Experimental Setup}

\subsection{Datasets}

\paragraph{Spider}
Spider \citep{spider} is a large-scale, complex, cross-domain text-to-SQL benchmark comprising 10,181 questions and 5,693 distinct queries across 200 databases, each with multiple tables. 
This benchmark requires the model to adapt to an unseen DB schema because different DBs are used for training and testing.  

\paragraph{BIRD}
BIRD \citep{bird} is a new large-scale, cross-domain text-to-SQL benchmark comprising 12,751 unique question-SQL pairs across 95 large real-world databases. Compared with Spider, BIRD comprises considerably more complex SQL queries with various SQL keywords (\texttt{LEFT JOIN}, \texttt{PARTITION BY}, etc.) and functions (\texttt{IIF}, \texttt{CASE}, \texttt{ROUND}, etc.) that are not included in Spider. In addition, BIRD requires reasoning using external knowledge (such as synonym knowledge and value illustrations) to generate accurate SQL queries.

\subsection{Evaluation Metrics}

\paragraph{Execution Accuracy (EX)}
EX indicates whether the SQL execution result generated by the model is identical to the gold SQL query. Because an SQL query can be written in various forms to produce the same result, evaluation metrics based on string matching significantly underestimate model performance. 

\paragraph{Valid Efficiency Score (VES)}
For the BIRD dataset, \citet{bird} proposed an additional evaluation metric called VES that measures the efficiency of a valid model-generated query based on the execution time. A query is considered invalid and assigned a score of zero if its execution result differs from that of the gold SQL.
Therefore, VES considers both the model accuracy and efficiency for the generated query.

\subsection{Implementation Details}
In all of our experiments, we used the \texttt{GPT-4 8K} as the LLM and \texttt{text-embedding-ada-002} as the sentence embedding model, which was accessed via Azure OpenAI API.
Additionally, we employed the FAISS \citep{faiss} library for the embedding similarity search.
In schema linking, we used $p_t=3$ prompts for table linking and $p_c=3$ prompts for column linking. To generate multiple candidate SQL queries, we use $p_q=5$ distinct prompts.
For each GPT API call, we used a temperature of 1.0 and generated $n$=20 responses.
In both the SQL-generation and MCS steps, we used $k$=20 question-SQL pairs as few-shot examples.
We executed all candidate SQL queries with a timeout of 180s and filtered out queries with a confidence score lower than the threshold $T$=0.2.

\begin{table*}
    \centering
    \begin{tabular}{lccc|c}
    \toprule
        Method & Simple & Moderate & Challenging & ALL \\
    \midrule
        zero-shot & 59.0 & 40.2 & 30.6 & 50.7 \\
        \quad + schema linking & 61.4 & 41.1 & 35.4 & 52.8 (+2.1) \\
        \quad + sample table contents & 64.1 & 42.6 & 38.9 & 55.2 (+2.4) \\
        \quad + few-shot examples & 67.1 & 51.6 & 41.0 & 60.0 (+4.8) \\
        \quad + MCS ($p_q$=1, $n$=20) & 69.3 & 52.0 & 48.6 & 62.1 (+2.1) \\
        \quad + MCS ($p_q$=5, $n$=20) & \textbf{70.4} & \textbf{53.1} & \textbf{51.4} & \textbf{63.4} (+1.3) \\
    \bottomrule
    \end{tabular}
    \caption{Execution accuracies for the ablation analysis on the BIRD dev set across various difficulty levels.}
    \label{table:bird-ablation}
\end{table*}

\begin{table*}
    \centering
    \begin{tabular}{lcccc|c}
    \toprule
        Method & Easy & Medium & Hard & Extra Hard & ALL \\
    \midrule
        zero-shot & 79.0 & 83.4 & 66.7 & 52.4 & 74.6 \\
        \quad + schema linking & 86.7 & 85.0 & 67.2 & 57.2 & 77.9 (+3.3) \\
        \quad + sample table contents & 89.9 & 88.6 & 69.0 & 59.0 & 80.9 (+3.0) \\
        \quad + few-shot examples & \textbf{94.0} & 90.4 & 86.2 & 66.3 & 86.7 (+5.8) \\
        \quad + MCS ($p_q$=1, $n$=20) & 93.1 & 92.6 & \textbf{88.5} & 72.2 & 88.8 (+2.1) \\
        \quad + MCS ($p_q$=5, $n$=20) & \textbf{94.0} & \textbf{93.5} & \textbf{88.5} & \textbf{72.9} & \textbf{89.5} (+0.7) \\
    \bottomrule
    \end{tabular}
    \caption{Execution accuracies for the ablation analysis on the Spider dev set across various difficulty levels.}
    \label{table:spider-ablation}
\end{table*}

\subsection{Baselines}

We compare the proposed MCS-SQL approach with ICL-based methods based on GPT-4.

\paragraph{GPT-4 \citep{gpt-4}}
uses the zero-shot prompt provided in OpenAI's text-to-SQL demo\footnote{\href{https://platform.openai.com/examples/default-sql-translate}{https://platform.openai.com/examples/default-sql-translate}}.

\paragraph{DIN-SQL \citep{din-sql}} classifies the complexity of the question and generates an SQL query by applying different prompts based on the classification result. In each step, it uses a prompt with fixed few-shot examples that are manually written in the CoT style.

\paragraph{DAIL-SQL \citep{dail-sql}} employs dynamic few-shot examples by considering the similarity of both the questions and the queries. For additional performance improvement, self-consistency \citep{self-consistency} is introduced.

\paragraph{MAC-SQL \citep{mac-sql}} decomposes the question into sub-questions and sequentially generates SQL queries for each sub-question using manually crafted few-shot samples. Additionally, in case of a syntax error, it uses a prompt to correct the generated query.

\section{Results and Analysis}

\subsection{Main Results}

\paragraph{BIRD}

Table~\ref{table:bird-main} presents the EX and VES of the proposed and baseline models for the BIRD dev and test sets. The results demonstrate that the proposed approach significantly outperforms existing ICL-based approaches in both metrics. Specifically, the proposed method achieved an EX of 65.45\% and a VES of 71.35\% on the holdout test set, surpassing the performance of the previous SOTA ICL-based approach \citep{mac-sql} by significant margins of 5.86\% and 3.67\%, respectively. Furthermore, the proposed method established a new SOTA performance on the BIRD, surpassing the former SOTA method with a substantial margin of 4.74\% in EX and 3.67\% in VES.

\paragraph{Spider}

Table~\ref{table:spider-main} presents the EX of the proposed and baseline methods for the Spider dev and test sets. Similar to the results obtained for BIRD, the proposed approach significantly outperforms all existing ICL-based approaches. Specifically, on the dev set, our approach achieved an EX of 89.5, which exceeds that of the former SOTA ICL-based approach \citep{mac-sql} by +2.7\%.

\subsection{Ablation Study}
We conducted an ablation study to investigate the incremental impact of each component of the proposed approach on the EX. The ablation results for the BIRD dev set are presented in Table~\ref{table:bird-ablation}.
The addition of schema linking to the baseline zero-shot setting resulted in a 2.1\% improvement. This underscores the importance of refining the schema prior to SQL generation and shows that the proposed schema-linking process effectively selects relevant tables and columns. The inclusion of the sample table contents in the prompt further amplified this gain by +2.4\%. The introduction of dynamic few-shot examples that were selected based on masked question similarity resulted in the largest performance improvement of +4.8\%. Moreover, when we sampled multiple answers from the LLM using the same prompt and employed the proposed MCS method, the performance further improved by 2.1\%. This demonstrates the capability of the proposed SQL selection method in discerning and selecting the most accurate query from a set of candidates. Finally, introducing multiple prompts led to further enhancements of +1.3\%, particularly showing significant performance improvement on challenging queries. This improvement demonstrates that broadening the search space using various prompts significantly boosted the SQL-generation accuracy.

Table~\ref{table:spider-ablation} lists the ablation results for the Spider dev set, wherein it is evident that each component of the proposed approach contributed to significant performance gains, similar to the results obtained for BIRD. This consistent performance enhancement across different benchmarks confirms the effectiveness and adaptability of the proposed approach for the text-to-SQL task.

\subsection{Impact of Using Multiple Prompts in Schema Linking}
\label{sec:effect-schema-linking}
We conducted a comparative analysis of the following three cases to investigate the impact of using multiple prompts in schema linking: (1) greedy decoding with a single prompt; (2) taking the union of multiple answers generated from a single prompt; and (3) taking the union of multiple answers generated from multiple prompts. Table~\ref{table:effect-schema-linking} lists the recall of schema linking for each case, which was calculated based on whether the predicted list of tables and columns included those used in the gold query. 

The results demonstrate that sampling multiple responses using the same prompt and aggregating them led to notable performance gains of +15.8\% for BIRD and +4.7\% for Spider. However, leveraging multiple prompts contributed to further significant improvements, with gains of +12.7\% for BIRD and +2.7\% for Spider. These results indicate that the order of tables and columns in the prompt affects the schema-linking results of the LLM and that the proposed multiple-prompt approach effectively mitigates this sensitivity. This improvement was particularly noticeable in BIRD, implying that the effectiveness of using multiple prompts increases for larger and more complex DB schemas.

As it is impossible to generate accurate SQL queries in subsequent processes if the necessary tables or columns are omitted in schema linking, our proposal of using the union of various responses from multiple prompts is pivotal for enhancing the SQL-generation performance.

\begin{table}[h]
\centering
\begin{tabular}{lcc}
    \toprule
        Method & BIRD & Spider \\
    \midrule
        $p_t=p_c=1$, $n$=1 & 61.3 & 91.8 \\
        $p_t=p_c=1$, $n$=20 & 77.1 & 96.5 \\
        \makecell{$p_t=p_c=3$, $n$=20 \\ (proposed)} & \textbf{89.8} & \textbf{99.2} \\
    \bottomrule
\end{tabular}
\caption{Recall of schema linking for the BIRD and Spider dev sets under three different settings: (1) greedy decoding with a single prompt, (2) sampling multiple answers from the same prompt, and (3) employing multiple prompts.}
\label{table:effect-schema-linking}
\end{table}

\subsection{Impacts of Different Few-shot Selection Strategies}
\label{sec:effect-few-shot}
Table~\ref{table:effect-few-shot} presents the EXs when different few-shot strategies were employed. 
The performance improved significantly by selecting few-shot examples based on question similarity instead of random selection, with an increase of 2.3\% for BIRD and 4.2\% for Spider.
Additionally, a further performance boost was noted by predicating the selection on the similarity of the masked question rather than the original question, with enhancements of 0.7\% and 0.5\% for BIRD and Spider, respectively.

\begin{table}[h]
\centering
\begin{tabular}{lcc}
    \toprule
        Method & BIRD & Spider \\
    \midrule
        Random & 57.0 & 82.0 \\
        Question Similarity & 59.3 & 86.2 \\
        Masked Question Similarity & \textbf{60.0} & \textbf{86.7} \\
    \bottomrule
\end{tabular}
\caption{Execution accuracies under different few-shot examples selection strategies for the BIRD and Spider dev sets. }
\label{table:effect-few-shot}
\end{table}

\subsection{Impact of MCS}
\label{sec:effect-mcq}

During the SQL selection phase (\cref{subsec:selection}), we explored whether the proposed MCS via LLM was more effective than a majority vote, which selects the query with the highest confidence score.
As presented in Table~\ref{table:effect-mcs}, the proposed MCS approach outperformed the majority vote approach by +0.6\% and +0.3\% for the BIRD and Spider, respectively. Notably, in the absence of confidence-based filtering, as expressed in Eq. (\ref{eq:confidence-threshold}), the efficacy of the MCS method decreased significantly. This result underscores the importance of employing confidence-based filtering to effectively narrow down the candidate pool when using MCS.

\begin{table}[h]
\centering
\begin{tabular}{lcc}
    \toprule
        Method & BIRD & Spider \\
    \midrule
        Majority Vote & 62.8 & 89.2 \\
        MCS w/o confidence filtering & 62.5 & 85.4 \\
        \makecell{MCS w/ confidence filtering \\ (proposed)} & \textbf{63.4} & \textbf{89.5} \\
    \bottomrule
\end{tabular}
\caption{Execution accuracies under different SQL selection strategies for the BIRD and Spider dev sets.}
\label{table:effect-mcs}
\end{table}

\section{Conclusion}

This study introduces a novel method that leverages multiple prompts to enhance the accuracy and robustness of ICL-based text-to-SQL generation. 
Specifically, the proposed approach performs robust schema linking using distinct prompts.
In addition, we employ different few-shot selection strategies to generate multiple query generation prompts, which yield various candidate SQL queries. These candidates are subsequently filtered based on their confidence scores, and the optimal query is selected using the LLM with MCS.
Evaluations on the BIRD and Spider benchmarks showed that the proposed approach significantly outperforms existing ICL-based approaches and achieved a new SOTA performance on the BIRD.







\bibliography{anthology,custom}
\bibliographystyle{acl_natbib}

\newpage
\appendix

\section{Error Analysis}

\begin{figure*}[b]
\centering\includegraphics[scale=0.2]{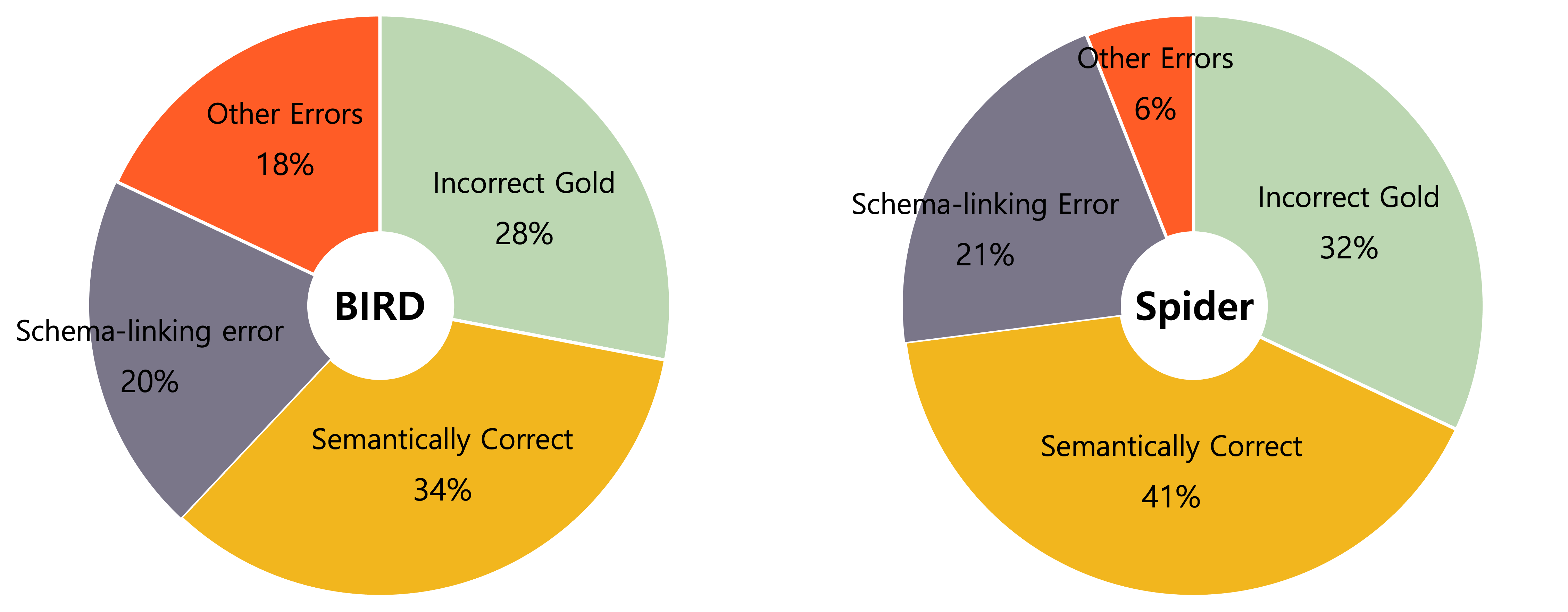}
\caption{\label{fig:error-analysis} Error analysis for the Spider and BIRD dev sets. }

\end{figure*}

We performed an error analysis to gain a deeper understanding of the instances in which the proposed method failed to accurately predict the SQL query. This analysis was conducted on a random sample of 100 examples from both the Spider and BIRD dev sets, where the execution results of the predicted query and gold SQL differed. Failure cases were categorized into four distinct categories as follows:
\begin{itemize}
    \item \textbf{Incorrect Gold}: The gold SQL query provided by human annotators in the dataset was incorrect.
    \item \textbf{Semantically Correct}:  The predicted SQL query was semantically equivalent to the gold SQL query, but the execution result differed owing to factors such as ties in the output \citep{evaluating}, the order of columns in the SELECT clause, or the inclusion of additional columns in the SELECT clause.
    \item \textbf{Schema Linking Error}: The predicted SQL query referred to different tables or columns than those referred to in the gold SQL query.
    \item \textbf{Other Errors}: The predicted SQL contained errors other than schema linking.
\end{itemize}
Examples from each category are listed in Table~\ref{table:error-analysis}.

The results of manually classifying the failure cases are depicted in Figure~\ref{fig:error-analysis}. 
The 62 and 73\% failure cases in the BIRD and Spider benchmarks, respectively, were those wherein the correct query was generated but they were still considered failures owing to inaccuracies in the gold queries or limitations of the method used to calculate the EX.
When these cases were excluded, the most prevalent error was schema linking, which involved selecting incorrect tables or columns (20\% for BIRD and 21\% for Spider).
However, even for human experts, referencing the exact tables and columns as the gold query is a challenging task because of inherent ambiguities. Such ambiguities arise when multiple columns in a DB share the same semantic meaning, or when the question does not explicitly specify the columns to be included in the SELECT clause.
Other errors, excluding schema linking, accounted for only 6\% and 18\% of the errors for Spider and BIRD, respectively. These errors encompassed cases in which the question or evidence was misinterpreted, incorrect assumptions were made regarding the DB content, or a syntactically correct SQL query could not be generated.

These results indicate that the EX of the proposed approach was \textbf{significantly underestimated} for both datasets. Furthermore, this analysis underscores the need for more precise gold queries and robust evaluation methodologies for text-to-SQL tasks.

\begin{table*}
\small
    \centering
    \begin{tabular}{p{3cm}p{12cm}}
    \toprule
        \textbf{Incorrect Gold} &  \\
    \midrule
        Question & List out the code for drivers who have nationality in America. \\
        Evidence & nationality = 'America' \\
        GOLD & SELECT code FROM drivers WHERE Nationality = \textcolor{blue}{'American'} \\
        PRED & SELECT code FROM drivers WHERE nationality = \textcolor{red}{'America'} \\
    \midrule
        Question & Which customer paid the most in 2012/8/25? \\
        Evidence & '2012/8/25' can be represented by '2012-08-25' \\
        GOLD & SELECT CustomerID FROM transactions\_1k WHERE Date = '2012-08-25' GROUP BY CustomerID ORDER BY \textcolor{blue}{SUM(Price)} DESC LIMIT 1 \\
        PRED & SELECT CustomerID FROM transactions\_1k WHERE Date = '2012-08-25' GROUP BY CustomerID ORDER BY \textcolor{red}{SUM(Amount * Price)} DESC LIMIT 1 \\
    \midrule
        \textbf{Semantically Correct} & \\
    \midrule
        Question & What is the phone number and extension number for the school with the zip code 95203-3704? Indicate the school's name. \\
        GOLD & SELECT \textcolor{blue}{Phone, Ext, School} FROM schools WHERE Zip = '95203-3704' \\
        PRED & SELECT \textcolor{red}{School, Phone, Ext} FROM schools WHERE Zip = '95203-3704' \\
    \midrule
        Question & List the card names with value that cost more converted mana for the face. \\
        Evidence & more converted mana for the face refers to Max(faceConvertedManaCost); \\
        GOLD & SELECT name FROM cards \textcolor{blue}{ORDER BY faceConvertedManaCost LIMIT 1} \\
        PRED & SELECT name FROM cards \textcolor{red}{WHERE faceConvertedManaCost = ( SELECT MAX(faceConvertedManaCost) FROM cards )} \\
    \midrule
        \textbf{Schema Linking Error} & \\
    \midrule
        Question & Which constructor has the highest point? \\
        GOLD & SELECT T2.name FROM \textcolor{blue}{constructorStandings} AS T1 INNER JOIN constructors AS T2 on T1.constructorId = T2.constructorId ORDER BY T1.points DESC LIMIT 1 \\
        PRED & SELECT T2.name FROM \textcolor{red}{constructorResults} AS T1 INNER JOIN constructors AS T2 ON T1.constructorId = T2.constructorId ORDER BY T1.points DESC LIMIT 1 \\
    \midrule
        Question & List all the mythic rarity print cards banned in gladiator format.  \\
        Evidence & mythic rarity printing refers to rarity = 'mythic'; card banned refers to status = 'Banned'; in gladiator format refers to format = 'gladiator'; \\
        GOLD & SELECT DISTINCT \textcolor{blue}{T1.id} FROM cards AS T1 INNER JOIN legalities AS T2 ON T1.uuid = T2.uuid WHERE T2.format = 'gladiator' AND T2.status = 'Banned' AND T1.rarity = 'mythic' \\
        PRED & SELECT DISTINCT \textcolor{red}{cards.name} FROM cards INNER JOIN legalities ON cards.uuid = legalities.uuid WHERE legalities.format = 'gladiator' AND legalities.status = 'Banned' AND cards.rarity = 'mythic' \\
   \midrule
     \textbf{Other Errors} & \\
   \midrule
     Question & What is the lowest grade for the District Special Education Consortia School with National Center for Educational Statistics school district identification number of 613360? \\
     Evidence & District Special Education Consortia School refers to EdOpsCode = 'SPECON'. \\
     GOLD & SELECT \textcolor{blue}{MIN(T1.`Low Grade`)} FROM frpm AS T1 INNER JOIN schools AS T2 ON T1.CDSCode = T2.CDSCode WHERE T2.NCESDist = 613360 AND T2.EdOpsCode = 'SPECON' \\
     PRED & SELECT \textcolor{red}{frpm.`Low Grade`} FROM frpm INNER JOIN schools ON frpm.CDSCode = schools.CDSCode WHERE schools.EdOpsCode = 'SPECON' AND schools.NCESDist = '613360' \\
   \midrule
     Question & What type of promotion is of card 'Duress'? \\
     Evidence & card Duress refers to name = 'Duress'; type of promotion refers to promoTypes; \\
     GOLD & SELECT promoTypes FROM cards WHERE name = 'Duress' \textcolor{blue}{AND promoTypes IS NOT NULL} \\
     PRED & SELECT promoTypes FROM cards WHERE name = 'Duress' \\
     
    \bottomrule
    \end{tabular}
    \caption{Examples of each error case in the BIRD dev set.}
    \label{table:error-analysis}
\end{table*}

\newpage
\onecolumn
\section{Prompts}
\label{sec:schema-linking-prompts}

\subsection{Prompt for Schema Linking}

\subsubsection{Prompt for Table Linking}
\label{appendix:table-linking-prompt}
\begin{lstlisting}{language={}, caption=An example of a full prompt for table linking., label={lst:full-table-linking}}
### Given a database schema, question, and knowledge evidence, extract a list of tables that should be referenced to convert the question into SQL.

### SQLite SQL tables, with their properties:
# molecule ( molecule_id, label )
# connected ( atom_id, atom_id2, bond_id )
# bond ( bond_id, molecule_id, bond_type )
# atom ( atom_id, molecule_id, element )
#
# atom.molecule_id = molecule.molecule_id
# bond.molecule_id = molecule.molecule_id
# connected.bond_id = bond.bond_id
# connected.atom_id2 = atom.atom_id
# connected.atom_id = atom.atom_id

### Question: Among all chemical compounds identified in the database, what percent of compounds form a triple-bond.
### Knowledge Evidence: triple bond refers to bond_type = '#';

You need to not only select the required tables, but also explain in detail why each table is needed.
Your answer should strictly follow the following json format.
{
  "reasoning": "",  // The reason for choosing each table.
  "tables": [],  // List of selected tables.
}

### Your Answer: 
\end{lstlisting}

\subsubsection{Prompt for Column Linking}
\label{appendix:column-linking-prompt}
\begin{lstlisting}{language={}, caption=An example of a full prompt for column linking., label={lst:full-table-linking}}
### Given a database schema, question, and knowledge evidence, extract a list of columns that should be referenced to convert the question into SQL.

### SQLite SQL tables, with their properties:
# molecule ( molecule_id, label )
# bond ( bond_id, molecule_id, bond_type )
#
# bond.molecule_id = molecule.molecule_id

### Question: Among all chemical compounds identified in the database, what percent of compounds form a triple-bond.
### Knowledge Evidence: triple bond refers to bond_type = '#';

You need to not only select the required columns, but also explain in detail why each column is needed.
Your answer should strictly follow the following json format.
{{
  "reasoning": "",  // The reason for choosing each column.
  "columns": ["table_name_i.column_name_j", ...],  // List of selected columns 
}}

### Your Answer:
\end{lstlisting}

\subsection{Prompt for Question Masking}
\label{appendix:question-masking-prompt}
\begin{lstlisting}[language={}, caption=An example of a full prompt for question masking., label={lst:full-question-masking}]
### Given a DB schema and a question, mask the table name, column name, and values in the question.

<example1>
### SQLite SQL tables, with their properties:
# customers ( CustomerID: integer, Segment: text, Currency: text )
# gasstations ( GasStationID: integer, ChainID: integer, Country: text, Segment: text )
# products ( ProductID: integer, Description: text )
# transactions_1k ( TransactionID: integer, Date: date, Time: text, CustomerID: integer, CardID: integer, GasStationID: integer, ProductID: integer, Amount: integer, Price: real )
# yearmonth ( CustomerID: integer, Date: text, Consumption: real )

### Question: For all the people who paid more than 29.00 per unit of product id No.5. Give their consumption status in the August of 2012.
### Masked Question: For all the [TABLE] who paid more than [VALUE] per unit of [COLUMN] [VALUE]. Give their consumption status in the [VALUE]. 
</example1>

<example2>
### SQLite SQL tables, with their properties:
# customers ( CustomerID: integer, Segment: text, Currency: text )
# gasstations ( GasStationID: integer, ChainID: integer, Country: text, Segment: text )
# products ( ProductID: integer, Description: text )
# transactions_1k ( TransactionID: integer, Date: date, Time: text, CustomerID: integer, CardID: integer, GasStationID: integer, ProductID: integer, Amount: integer, Price: real )
# yearmonth ( CustomerID: integer, Date: text, Consumption: real )

### Question: How much did customer 6 consume in total between August and November 2013?
### Masked Question: How much did [TABLE] [VALUE] consume in total between [VALUE] and [VALUE]?
</example2>

<example3>
### SQLite SQL tables, with their properties:
# drivers ( driverId: integer, driverRef: text, number: integer, code: text, forename: text, surname: text, dob: date, nationality: text, url: text )

### Question: How many Australian drivers who were born in 1980? 
### Masked Question: How many [VALUE] [TABLE] who were born in [VALUE]?
</example3>


### SQLite SQL tables, with their properties:
# molecule ( molecule_id, label )
# bond ( bond_id, molecule_id, bond_type )
#
# bond.molecule_id = molecule.molecule_id

### Question: Among all chemical compounds identified in the database, what percent of compounds form a triple-bond.
### Knowledge Evidence: triple bond refers to bond_type = '#';

### Masked Question: 
\end{lstlisting}

\subsection{Prompt for SQL Generation}
\label{appendix:sql-generation-prompt}
\begin{lstlisting}{language={}, caption=An example of a full prompt for SQL generation., label={lst:full-sql-generation}}
### Given a database schema, question, and knowledge evidence, generate the correct sqlite SQL query for the question.

<examples>
# Question: Among all the customers, what is the percentage of the customer's nation being Germany?
# Knowledge Evidence: DIVIDE(COUNT(c_custkey when n_name = 'GERMANY'), COUNT(c_custkey)) as percentage;
# Gold SQL: SELECT CAST(SUM(IIF(T2.n_name = 'GERMANY', 1, 0)) AS REAL) * 100 / COUNT(T1.c_custkey) FROM customer AS T1 INNER JOIN nation AS T2 ON T1.c_nationkey = T2.n_nationkey

# Question: Among the schools whose donators are teachers, what is the percentage of schools that are in Brooklyn?
# Knowledge Evidence: donors are teachers refers to is_teacher_acct = 't'; Brooklyn is school_city; percentage = Divide(Count(school_city-'Brooklyn'),Count(school_city))*100
# Gold SQL: SELECT CAST(SUM(CASE WHEN T1.school_city LIKE 'Brooklyn' THEN 1 ELSE 0 END) AS REAL) * 100 / COUNT(T1.teacher_acctid) FROM projects AS T1 INNER JOIN donations AS T2 ON T1.projectid = T2.projectid WHERE T2.is_teacher_acct = 't'

...
</examples>


### SQLite SQL tables, with their properties:
# molecule ( molecule_id, label )
# bond ( bond_id, molecule_id, bond_type )
#
# bond.molecule_id = molecule.molecule_id

### The type and description of each column:
# [molecule]
- molecule_id (text): unique id of molecule
- label (text): whether this molecule is carcinogenic or not

# [bond]
- bond_id (text): unique id representing bonds
- molecule_id (text): identifying the molecule in which the bond appears
- bond_type (text): type of the bond

### Sample rows of each table in csv format:
# [molecule]
molecule_id,label
TR000,+
TR001,+
TR002,-

# [bond]
bond_id,molecule_id,bond_type
TR000_1_2,TR000,-
TR000_2_3,TR000,-
TR000_2_4,TR000,-



### Question: Among all chemical compounds identified in the database, what percent of compounds form a triple-bond.
### Knowledge Evidence: triple bond refers to bond_type = '#';

You need to not only create the SQL, but also provide the detailed reasoning steps required to create the SQL. Your answer should strictly follow the following json format:
{
  "reasoning": "",  // The reasoning steps for generating SQL.
  "sql": "",  // The final generated SQL.
}

### Your Answer:
\end{lstlisting}

\subsection{Prompt for SQL Selection}
\label{appendix:mcs-prompt}
\begin{lstlisting}{language={}, caption=An example of a full prompt for SQL selection., label={lst:full-sql-selection}}
### When a DB schema, a question, and a knowledge evidence are given, and up to three SQLite queries expressing the question are given, please choose the most accurate SQL based on the Checklist.

<examples>
# Question: Among all the customers, what is the percentage of the customer's nation being Germany?
# Knowledge Evidence: DIVIDE(COUNT(c_custkey when n_name = 'GERMANY'), COUNT(c_custkey)) as percentage;
# Gold SQL: SELECT CAST(SUM(IIF(T2.n_name = 'GERMANY', 1, 0)) AS REAL) * 100 / COUNT(T1.c_custkey) FROM customer AS T1 INNER JOIN nation AS T2 ON T1.c_nationkey = T2.n_nationkey

# Question: Among the schools whose donators are teachers, what is the percentage of schools that are in Brooklyn?
# Knowledge Evidence: donors are teachers refers to is_teacher_acct = 't'; Brooklyn is school_city; percentage = Divide(Count(school_city-'Brooklyn'),Count(school_city))*100
# Gold SQL: SELECT CAST(SUM(CASE WHEN T1.school_city LIKE 'Brooklyn' THEN 1 ELSE 0 END) AS REAL) * 100 / COUNT(T1.teacher_acctid) FROM projects AS T1 INNER JOIN donations AS T2 ON T1.projectid = T2.projectid WHERE T2.is_teacher_acct = 't'

...
</examples>


### SQLite SQL tables, with their properties:
# molecule ( molecule_id, label )
# bond ( bond_id, molecule_id, bond_type )
#
# bond.molecule_id = molecule.molecule_id

### The type and description of each column:
# [molecule]
- molecule_id (text): unique id of molecule
- label (text): whether this molecule is carcinogenic or not

# [bond]
- bond_id (text): unique id representing bonds
- molecule_id (text): identifying the molecule in which the bond appears
- bond_type (text): type of the bond

### Sample rows of each table in csv format:
# [molecule]
molecule_id,label
TR000,+
TR001,+
TR002,-

# [bond]
bond_id,molecule_id,bond_type
TR000_1_2,TR000,-
TR000_2_3,TR000,-
TR000_2_4,TR000,-



### Question: Among all chemical compounds identified in the database, what percent of compounds form a triple-bond.
### Knowledge Evidence: triple bond refers to bond_type = '#';

### Candidate SQLs:
1. SELECT CAST(COUNT(CASE WHEN bond_type = '#' THEN 1 ELSE NULL END) AS REAL) * 100 / COUNT(*) FROM bond
2. SELECT CAST(COUNT(DISTINCT CASE WHEN bond_type = '#' THEN molecule_id ELSE NULL END) AS REAL) * 100 / COUNT(DISTINCT molecule_id) FROM bond

### Checklist:
1. The SQL should accurately represent the question.
2. The SQL should accurately use the given knowledge evidence.
3. The SELECT clause should not include any additional columns that are not included in the question.
4. The order of columns in the SELECT clause must be the same as the order in the question.
5. Check if the operations are being performed correctly according to the column type.

### Instruction:
- If the first SQL satisfies all the conditions of the checklist, please choose the first SQL. If not, move on to the next SQL.
- If there's no SQL that satisfies all the requirements on the checklist, just choose the first SQL.
- Provide a detailed step-by-step explanation following the order of the checklist when checking whether each SQL satisfies the checklist.
- Your answer should strictly follow the following json format.
{{
  "reasoning": "",  // The reasoning steps for choosing the best SQL.
  "sql": "",  // The final chosen SQL.
}}

### Your Answer:
\end{lstlisting}

\end{document}